\documentclass[letterpaper]{article} 
\usepackage[]{aaai2026}
\usepackage{times}  
\usepackage{helvet}  
\usepackage{courier}  
\usepackage[hyphens]{url}  
\usepackage{graphicx} 
\usepackage{multirow}
\usepackage{booktabs}
\usepackage{xcolor}
\usepackage{natbib}  
\usepackage{caption} 
\usepackage{colortbl}
\usepackage{amsmath}
\usepackage{fvextra}
\DefineVerbatimEnvironment{VerbatimWrap}{Verbatim}{
breaklines=true, 
breakindent=0pt, 
breaksymbol={},
fontsize=\small,
}

\usepackage{algorithm}
\usepackage{algorithmic}
\usepackage{newfloat}
\usepackage{listings}
\DeclareCaptionStyle{ruled}{labelfont=normalfont,labelsep=colon,strut=off} 
\floatstyle{ruled}
\newfloat{listing}{tb}{lst}{}
\floatname{listing}{Listing}
\usepackage{natbib}
\usepackage{csquotes}
\usepackage{url}
\usepackage{svg}
\usepackage{bm}
\usepackage{graphics}
\usepackage{graphicx}
\usepackage{array}
\usepackage[english]{babel}
\usepackage{textcomp}
\usepackage{latexsym}
\usepackage{siunitx}  
\usepackage{titletoc}

\usepackage{tikz}
\usetikzlibrary{tikzmark}
\usepackage[most]{tcolorbox}

\definecolor{myred}{rgb}{0.7, 0.3, 0.0}
\definecolor{myblue}{rgb}{0.2, 0.3, 0.6}
\definecolor{mygreen}{HTML}{008000}

\newcommand{\ours}{HiRA}
\title{HiRA: A Hierarchical Reasoning Framework for Decoupled Planning and Execution in Deep Search}
\author{%
Jiajie Jin$^{1}$, Xiaoxi Li$^{1}$,Yuyao Zhang$^{1}$, Guanting Dong$^{1}$, Yutao Zhu$^{1}$, \\ Zhao Yang$^{1}$, Hongjin Qian$^{2}$, Zhicheng Dou$^{1}$\thanks{Corresponding author.}
} 
\affiliations{
    $^1$Gaoling School of Artificial Intelligence, Renmin University of China \\
    $^2$Beijing Academy of Artificial Intelligence

    \{jinjiajie, dou\}@ruc.edu.cn
}

\usepackage{bibentry}

\begin{document}

\maketitle

\begin{abstract}
Complex information needs in real-world search scenarios demand deep reasoning and knowledge synthesis across diverse sources, which traditional retrieval-augmented generation (RAG) pipelines struggle to address effectively. Current reasoning-based approaches suffer from a fundamental limitation: they employ a single model to handle both high-level planning and detailed execution, resulting in inefficient reasoning and limited scalability. 
In this paper, we introduce \ours{}, a hierarchical framework that separates strategic planning from specialized execution. Our approach decomposes complex search tasks into multiple subtasks, assigns each subtask to a domain-specific agent equipped with external tools and reasoning capabilities, and coordinates the results through a structured integration mechanism. 
This separation prevents execution details from disrupting high-level reasoning while enabling the system to leverage specialized expertise for different types of information processing.
Experiments on four complex, cross-modal deep search benchmarks show that \ours{} significantly outperforms state-of-the-art RAG and agent-based systems, highlighting the effectiveness of decoupled planning and execution for multi-step information seeking tasks.
\end{abstract}

\section{Introduction}
The information explosion on the Internet has made it increasingly difficult to find answers to complex queries, leading to the rapid development of deep search tasks that require understanding complex information needs and synthesizing accurate answers from multiple sources~\cite{xu2025comprehensivesurveydeepresearch}. However, traditional search engines only return ranked web pages based on keyword matching, requiring users to filter and collect information manually. 

While large language models (LLMs) equipped with web search can provide direct answers, they typically utilize direct information from search results, lacking deep reasoning and comprehensive analysis capabilities~\cite{rag, ragsurvey}. This has motivated the development of specialized AI agents for deep search tasks, such as OpenAI DeepSearch and Grok DeepSearch~\cite{grok_deepsearch}, which aim to bridge the gap between simple search-enhanced models and deep information seeking systems.

\begin{figure}[!t]
\centering
\includegraphics[width=\linewidth]{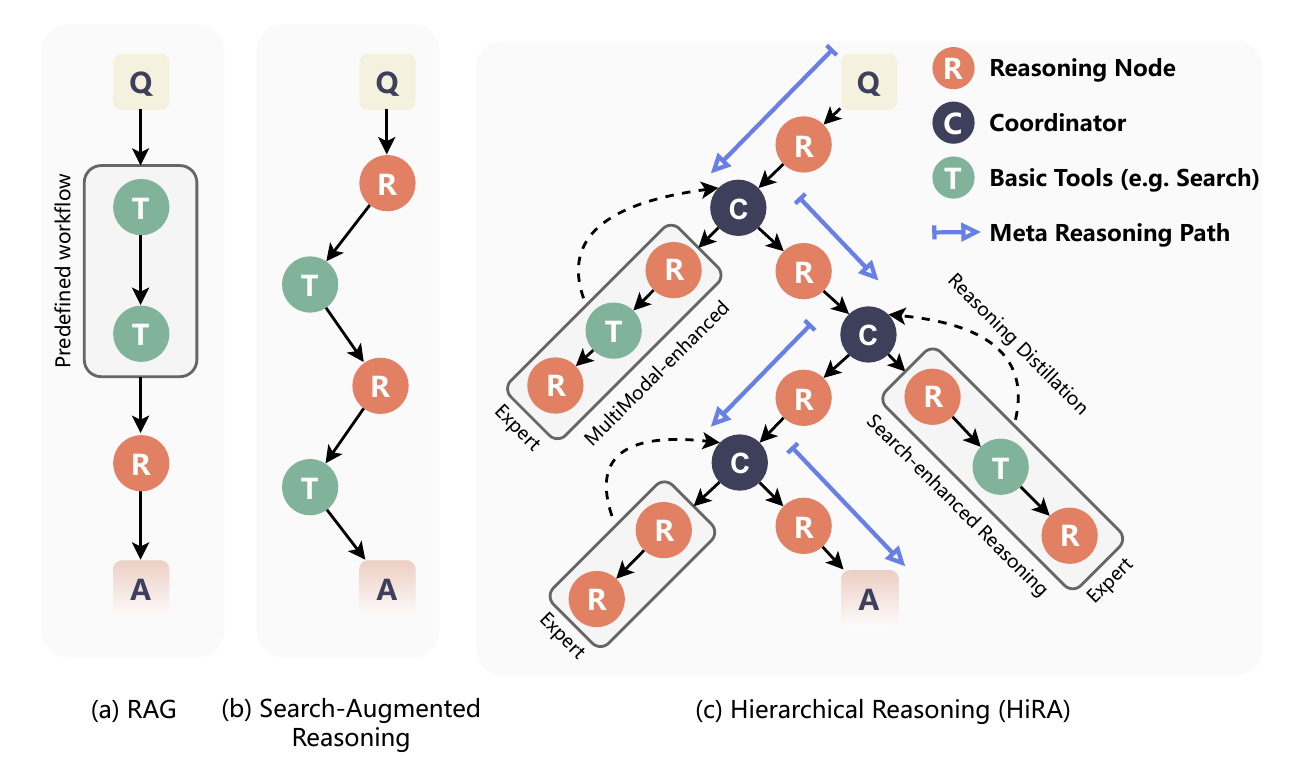}
\caption{Comparison of current approaches for deep search tasks: (a) Direct Reasoning with LRMs, (b) Search-Augmented Reasoning that enables LRMs to use a search engine during reasoning, and (c) Hierarchical Reasoning that autonomously interacts with expert agents and tools in a continuous thinking process.}
\label{fig:intro_fig}
\end{figure}

Conventional approaches typically employ retrieval-augmented generation (RAG) techniques with predefined workflows~\cite{rag, ircot, iterretgen}, which incorporate components like query decomposition, document summarization, and self-reflection to improve generation quality.
Recently, large reasoning models (LRMs) such as OpenAI-o1~\cite{openai2024reasoning} and DeepSeek-R1~\cite{deepseek-r1} have introduced new opportunities by integrating web search and browsing capabilities within their reasoning processes~\cite{searcho1, jin2025searchr1, song2025r1searcher}. As shown in Figure~\ref{fig:intro_fig}(b), these search-augmented reasoning methods can autonomously plan and acquire external knowledge for complex information retrieval in an end-to-end manner, significantly improving deep search performance. 

However, existing approaches suffer from architectural limitations due to their reliance on a single reasoning model to handle all tasks. Current methods typically work by prompting reasoning models to generate special tokens~\cite{searcho1, jin2025searchr1,Li2025WebThinker}, which are used to trigger corresponding tool activations during their thinking process. For example, in a recent method WebThinker~\cite{Li2025WebThinker}, the search action is triggered by \texttt{<|begin\_search\_query|>}.
This monolithic paradigm introduces two critical deficiencies:
(1) \textit{Limited capability extensibility}: Adding new tools or capabilities requires carefully redesigning prompts to teach the model how to use new token patterns and their application contexts. This process is brittle and often requires re-engineering token systems or extensive training to ensure reliable token generation and tool coordination.
(2) \textit{Reasoning disruption}: As shown in Figure~\ref{fig:intro_fig}, external execution results are directly injected into the main reasoning chain, introducing noise that may disturb core reasoning processes. This disruption weakens the model's logical thinking and occupies limited context windows with irrelevant operational details.
These limitations stem from the assumption that one agent is required to handle all aspects of complex reasoning tasks. We argue that an effective agent execution should follow a hierarchical structure, as shown in Figure~\ref{fig:intro_fig}(c): including a meta-agent for high-level planning, a coordinator for task reasoning transfer, and several specialized execution agents for specific operations. 
Each execution agent only needs to complete a single subtask through internal multi-step reasoning and iterative tool usage, allowing for deeper analysis without contaminating the overall planning process.

Based on this insight, we propose a \textbf{\underline{Hi}}erarchical  \textbf{\underline{R}}e\textbf{\underline{A}}soning (\ours{}) model, a framework designed to enhance deep search effectiveness by separating planning from execution. 
This architecture consists of three components: a meta reasoning planner, an adaptive reasoning coordinator, and a group of domain-specialized executors.
The meta reasoning planner breaks down complex tasks into subtasks through a reasoning process. These subtasks are then assigned to specialized agents by the adaptive reasoning coordinator. The assignments are based on the task complexity and required expertise. Each domain-specialized executor leverages specific reasoning models and external tools to execute the assigned subtask, with reasoning result distillation into the planner's process via the coordinator.

Our hierarchical design effectively decouples the strategic planning from the execution details, allowing for scalable, coherent reasoning in complex tasks. By integrating specialized expertise at the execution level and maintaining a coordinated flow across the hierarchy, \ours{} ensures both the flexibility and efficiency necessary for tackling advanced reasoning challenges. We conduct experiments on four complex, cross-modal, multi-scenario deep search tasks. The results demonstrate that our framework significantly outperforms existing methods across several aspects. 
Overall, the key contributions of this work are threefold: 

\begin{enumerate}
    \item \textbf{Hierarchical Reasoning Architecture:} We propose a novel hierarchical reasoning framework that integrates specialized tool-augmented reasoning agents as modules, eliminating the need for external tool orchestration or rigid predefined pipelines used in existing methods.
    
    \item \textbf{Enhanced Capability Integration:} The domain-specialized executors enable plug-and-play integration of diverse reasoning capabilities and tools. Existing search agents can be directly incorporated without prompt engineering or model re-training, preserving established workflows while enhancing performance.
    
    \item \textbf{Superior Empirical Performance:} Experiments across four complex cross-modal search tasks demonstrate significant improvements over traditional RAG and current agent-based methods, validating the effectiveness of hierarchical reasoning augmentation for complex information retrieval.
\end{enumerate}

\section{Related Work}
\paragraph{From Retrieval-Augmented Generation to Deep Search}
Retrieval-Augmented Generation (RAG) combines external knowledge with LLMs' parametric knowledge~\cite{rag, retro}, evolving from single-step retrieval~\cite{ragsurvey, flashrag_jiajie} to iterative pipelines with query decomposition~\cite{rqrag,querycompiler,ircot}, document refinement~\cite{recomp,bider_jiajie}, and multi-round search~\cite{iterretgen, ircot}. However, RAG methods rely on predefined workflows limiting adaptive decision-making. Recent LRMs integrate retrieval into reasoning~\cite{searcho1, jin2025searchr1} but still require inserting documents into reasoning chains or auxiliary summarization models~\cite{Li2025WebThinker}. These limitations motivate our hierarchical reasoning augmentation with specialized cognitive agents.
\paragraph{Planning-Execution Separation Approaches}
Recent work separates planning from execution to address information overload~\cite{bilal2025metathinking_survey}. \textit{Action-level separation} assigns executors to single-step tasks like Plan-Act~\cite{erdogan2025planandactimprovingplanningagents} and CoAct~\cite{hou2024coact}. \textit{Query-level separation} decomposes problems at higher granularity: REMA~\cite{wan2025rema} uses RL-based planners for mathematical reasoning, while LLMCompiler~\cite{kim2023llmcompiler} and Query Compiler~\cite{querycompiler} create parallel execution graphs. However, these methods suffer from rigid task decomposition and limited executor specialization beyond prompt variations. Our work addresses these limitations through dynamic reasoning delegation and domain-specialized agents within a hierarchical framework.

\section{Methodology}

\begin{figure*}[!t]
    \centering
    \includegraphics[width=0.95\linewidth]{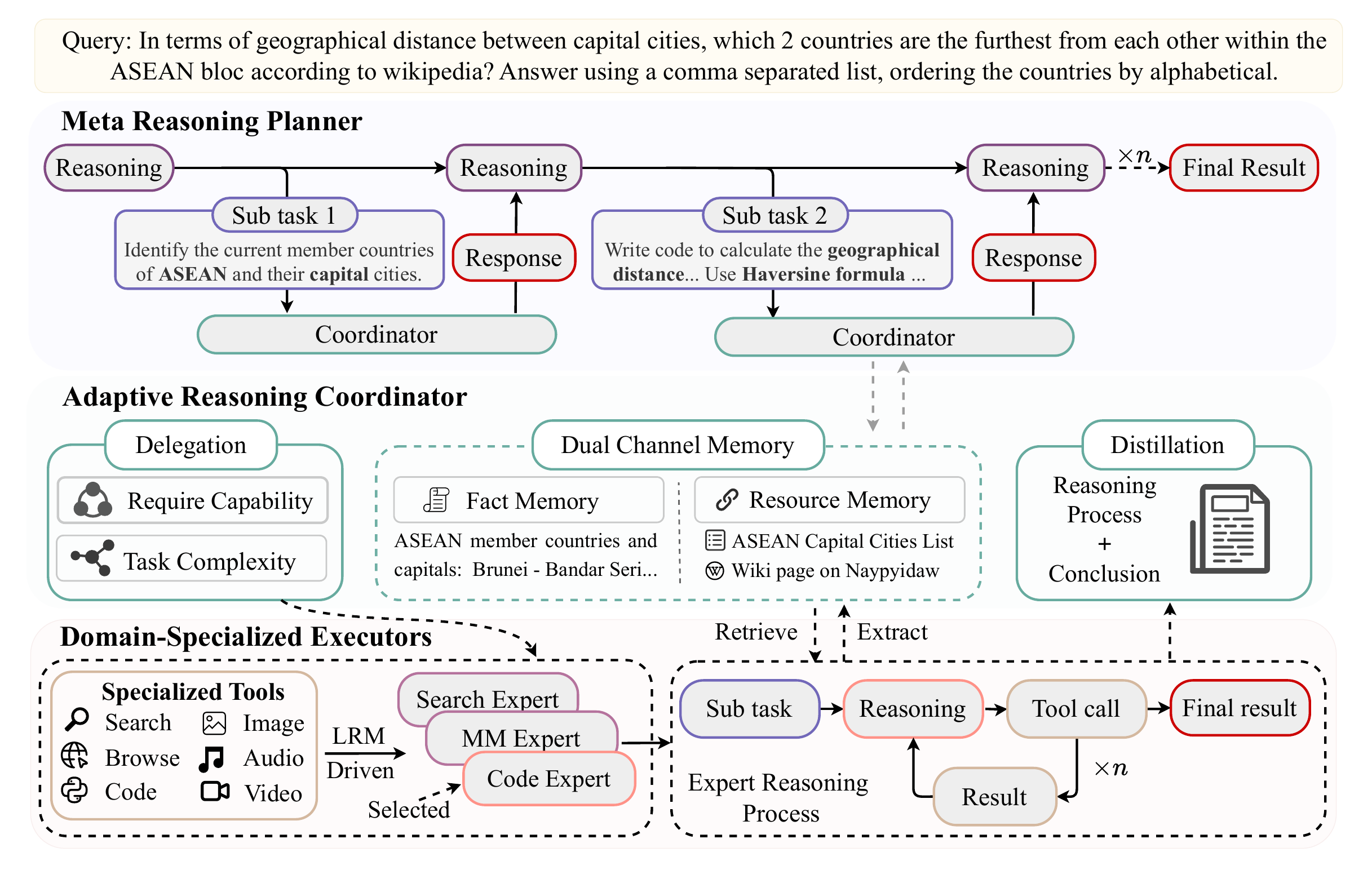}
    \caption{
    Overview of the \ours{} Framework.
    }
    \label{fig:overview}
\end{figure*}

\subsection{Problem Formulation}
Given a complex question $q$ that required information seeking and a predefined external environment $\mathcal{E}$, our objective is to design a framework that generates a final solution containing an answer $\mathcal{A}$ and the corresponding reasoning process $\mathcal{R}$. The generation process can be represented as:
\begin{equation*}
P(\mathcal{R}, a \mid q, \mathcal{E}) = \prod_{t=1}^{T_\mathcal{R}} P(\mathcal{R}_{t} \mid \mathcal{R}_{<t},q, \mathcal{E}_{< t}) \cdot P(a \mid q, \mathcal{R}),
\end{equation*}
where $T_\mathcal{R}$ represents the token generation steps for the reasoning process, $\mathcal{A}$ denotes the final answer, and $\mathcal{E}_{<t}={\{\mathcal{E}(R_{<s})\}}_{s<t}$ denotes the collection of all environment interaction results prior to timestep $t$.
While existing agentic reasoning approaches typically use tools directly as the external environment, our work introduces a higher-level abstraction where the environment $\mathcal{E}$ consists of a collection of expert agents, each capable of reasoning and utilizing specific tools to accomplish specialized tasks.

\subsection{Overview of the \ours{} Framework}

The \ours{} framework is a hierarchical reasoning system that enhances deep search effectiveness through the separation of planning and execution. As shown in Figure~\ref{fig:overview}, the framework consists of three modules: (1) a \textbf{meta reasoning planner} that decomposes complex tasks into subtasks through step-by-step reasoning, (2) an \textbf{adaptive reasoning coordinator} that assigns subtasks to appropriate domain-specialized executors (expert agents) based on task complexity and capabilities, and (3) a \textbf{domain-specialized executors} that execute subtasks using specialized reasoning models and external tools. Results obtained by the executors are integrated back into the planner's reasoning process through the coordinator. This hierarchical design decouples strategic planning from execution details, enabling scalable and coherent reasoning for complex information seeking tasks. 

In our framework, the meta reasoning planner operates at a high level of abstraction, decomposing the overarching task into high-level subtasks while deliberately omitting low-level implementation details. This design prevents the reasoning process from being biased or disrupted by fine-grained subtask specifics, thus enhancing the robustness and generalizability of the generated plans. A subtask is defined as a comprehensive, self-contained instruction that includes both the objective and the contextual requirements for accomplishing a specific component of the overall task. Unlike simple queries that typically request single pieces of information, subtasks are composite directives that may involve multiple steps of reasoning or tool calling, specify particular solution ideas, and include detailed requirements for information processing and result formatting. 
%

\subsection{Hierarchical Reasoning Paradigm}
\paragraph{$\bullet$ Meta Reasoning Planner}
The meta reasoning planner serves as the core module of the framework, responsible for planning, reasoning, and answer generation. Unlike conventional tool-augmented approaches that require models to directly invoke tools with specific parameters, our meta planner decouples the task into several high-level subtasks containing strategic instructions for expert agents. This design enables natural collaboration between meta and expert agents, ensuring smooth information transfer while eliminating the noise and overhead associated with direct tool invocation and execution-level decision making.

To enable dynamic subtask generation, we design a meta planner prompt that instructs the model to use special tokens for subtask dispatch. The overview of the process is shown in Figure~\ref{fig:overview}. In the reasoning process, the planner will automatically generate special tokens and place the description and requirements of subtasks in the middle, which is similar to the process of humans issuing tasks. As shown in Equation~(\ref{eq:subtask_generation}), the generated subtask description $s_k$ is based on previous task execution results $\{\mathcal{E}(s_j)\}_{j<k}$ and the current reasoning progress $\mathcal{O}_{<t}$, naturally enabling reflection, correction, supplementation, and continuity generation of prior tasks. We impose no explicit constraints on subtask scope, difficulty, size, or relationships with previous subtasks to preserve overall reasoning flexibility.

\begin{equation}
\label{eq:subtask_generation}
    P_{M}(s_k) = P_M(s_k \mid q,\mathcal{O}_{<t}, \{\mathcal{E}(s_j)\}_{j<k}).
\end{equation}

Then, the coordinator layer assigns $s_k$ to corresponding expert agents for execution. Execution results $\mathcal{A}_k(s_k)$ are wrapped in special tokens, and then integrated into the model's reasoning process for continued generation. Notably, the subtask execution results incorporated into the reasoning process contain essential execution procedures and final conclusions, rather than vanilla tool invocation results that may contain noise and require subsequent processing.

During the generation process, the model incrementally conditions on the original query $q$, the prior decoding history $\mathcal{O}_{<t}$, and the set of executed subtask results ${\mathcal{A}_j(s_j)}_{j \leq K}$ to derive the final answer $a$, formalized as:
\begin{equation}
\label{eq:final_answer}
P_{M}(a) = P_M(a \mid q, \mathcal{O}_{<t}, {\mathcal{A}_j(s_j)}_{j\leq K}),
\end{equation}
where $K$ denotes the total number of subtasks.

This design enables modular task planning by decoupling high-level goals from execution, allowing the model to generate subtasks without knowing specific expert agents.

\paragraph{$\bullet$ Adaptive Reasoning Coordinator} 
While separating execution from planning provides clear scalability advantages and reduces computational noise, it introduces the risk of information loss between components. To mitigate this challenge, we design an adaptive reasoning coordinator that incorporates bidirectional, task-specific reasoning transfer and a dual-channel memory mechanism. This coordinator facilitates seamless reasoning delegation from the meta agent to expert agents while enabling reasoning distillation in the reverse direction, thus preserving efficient inter-agent communication and maintaining the architectural benefits of separation.
The coordinator has three core functions as follows.

(1)~\textbf{Reasoning transfer process.} The coordinator is designed to interpret subtasks provided by the meta planner and identify the most suitable expert agent for task execution. Given the current subtask $s_k$ and detailed information about all experts $I_\mathcal{E} = \{\mathcal{I}_\mathcal{A}\}_{\mathcal{A} \in \mathcal{E}}$, the coordinator first analyzes the subtask requirements, then evaluates agent capabilities across two key dimensions before making the optimal selection. 
Our instruction framework $\mathcal{I}_{\text{select}}$ encompasses: (a)~Required capabilities: the domain knowledge and tool utilization abilities necessary for task completion, and (b)~Task complexity: the computational difficulty of the subtask and the required depth of analysis. For tasks within the same category (e.g., information retrieval), target information often resides at varying depths across data sources. While deploying the most sophisticated agent can ensure problem resolution, it may introduce unnecessary computational overhead and analytical redundancy. Therefore, the coordinator prioritizes selecting the most efficient agent for each specific task to optimize overall system performance.

Formally, we model this selection process as a classification problem:
\begin{equation*}
	\mathcal{A}_k^* = \text{argmax}_{\mathcal{A} \in \mathcal{E}} P_C(\mathcal{O}_{\text{dele}}^{(k)}, \mathcal{A} \mid s_k, \mathcal{I}_\mathcal{E}, \mathcal{I}_{\text{select}}),
\end{equation*}
where the coordinator generates selection reasoning $\mathcal{O}_{\text{dele}}^{(k)}$ and identifies the optimal expert agent $\mathcal{A}_k^*$ through chain-of-thought reasoning.

(2)~\textbf{Reasoning distillation process.} The coordinator is responsible for understanding and refining the expert agent's reasoning process before integrating the results into the meta reasoning planner's cognitive flow to maintain overall reasoning coherence. Unlike traditional tool-augmented reasoning approaches that primarily focus on final execution outputs, our framework considers both the expert agent's intermediate reasoning steps and the final conclusions. This dual consideration enables the meta planner to comprehend the underlying reasoning logic without being encumbered by low-level execution details, thereby facilitating autonomous reflection and critical evaluation of execution outcomes.

Specifically, given subtask $s_k$ and the expert agent's reasoning process $\mathcal{O}_{\text{expert}}^{(k)}$, the distilled reasoning process $\mathcal{O}_{\text{dist}}^{(k)}$ and refined conclusion $\mathcal{R}_{\text{dist}}^{(k)}$ are generated through:
\begin{equation*}
\begin{split}
&P_C\left(\mathcal{O}_{\text{dist}}^{(k)}, \mathcal{R}_{\text{dist}}^{(k)} \mid s_k, \mathcal{O}_{\text{expert}}^{(k)}\right) \\
&\quad = \underbrace{P_C\left(\mathcal{O}_{\text{dist}}^{(k)} \mid \mathcal{O}_{\text{expert}}^{(k)},\cdot \right)}_{\text{Reasoning Refinement}} \cdot \underbrace{P_C\left(\mathcal{R}_{\text{dist}}^{(k)} \mid \mathcal{O}_{\text{dist}}^{(k)},\mathcal{O}_{\text{expert}}^{(k)},\cdot \right)}_{\text{Conclusion Extraction}}.
\end{split}
\end{equation*}
Subsequently, $\mathcal{O}_{\text{dist}}^{(k)}$ and $\mathcal{R}_{\text{dist}}^{(k)}$ are concatenated and seamlessly integrated as reasoning distillation feedback into the meta reasoning planner's ongoing cognitive process.

(3)~\textbf{Dual-channel memory mechanism.} To enable effective information sharing and knowledge transfer among expert agents, we design a dual-channel memory mechanism tailored specifically for deep search task scenarios. The comprehensive memory repository $\mathcal{M}$ encompasses two distinct types of memory: fact memory $\mathcal{M}_f$ and resource memory $\mathcal{M}_r$.
Fact memory archives factual discoveries and insights extracted from expert agents' reasoning processes. Each memory entry comprises a factual assertion paired with its corresponding source attribution (URL, file name, or webpage identifier) to ensure traceability and verification. To maintain memory efficiency and reduce redundancy, multiple similar factual statements originating from identical sources are intelligently aggregated.
Resource memory maintains a repository of informational resources encountered during expert agent execution processes. Each entry contains a descriptive summary alongside the corresponding access path (e.g., webpage URL, file path), designed to provide subsequent agents with valuable exploration insights from previous agent interactions, thereby preventing redundant exploration and enhancing overall system efficiency.

Memory construction and utilization operate in conjunction with the reasoning transfer pipeline. During the reasoning distillation phase, relevant memory items are extracted from the reasoning process $\mathcal{O}_k$ and systematically updated within the global memory repository $\mathcal{M}$. During reasoning delegation, the coordinator intelligently retrieves pertinent memory entries based on semantic relevance between memory content and subtask requirements, as well as memory quality metrics, subsequently providing this contextual information as supplementary guidance to expert agents.

\begin{table*}[!t]
\centering
\small
\setlength\tabcolsep{3.2pt}
\begin{tabular}{p{3cm}*{4}{>{\centering\arraybackslash}p{1cm}}*{4}{>{\centering\arraybackslash}p{0.75cm}}*{4}{>{\centering\arraybackslash}p{0.75cm}}*{1}{>{\centering\arraybackslash}p{1.3cm}}}
\toprule
\multirow{2}[2]{*}{\textbf{Method}} & \multicolumn{4}{c}{\textbf{General AI Assistant}} & \multicolumn{4}{c}{\textbf{WebWalkerQA}} & \multicolumn{4}{c}{\textbf{Humanity's Last Exam}} & \textbf{SimpleQA} \\
\cmidrule(lr){2-5} \cmidrule(lr){6-9} \cmidrule(lr){10-13} \cmidrule(lr){14-14}
    & Level 1 & Level 2 & Level 3 & Avg. & Easy & Med. & Hard & Avg. & NS & CE & SF & Avg. & Acc \\
\midrule
\multicolumn{14}{l}{\textit{\textbf{Direct Reasoning}}} \\
Qwen3-32B-no-thinking & 14.3 & 6.1 & \underline{10.5} & 9.5 & 4.4 & 2.1 & 0.8 & 2.8 & 7.1 & 6.0 & 3.1 & 6.2 & 6.5 \\
Qwen3-32B-thinking & 26.2 & 12.1 & 0 & 14.9 & 6.9 & 1.1 & 2.9 & 3.1 & \underline{14.6} & \underline{9.8} & 8.4 & 12.6 & 10.5 \\
DeepSeek-R1-32B & 21.5 & 13.6 & 0.0 & 14.2 & 7.5 & 1.4 & 4.2 & 3.8 & 6.6 & 5.1 & 6.5 & 6.4 & 5.5 \\
QwQ-32B & 30.9 & 6.5 & 5.2 & 18.9 & 7.5 & 2.1 & 4.6 & 4.3 & 11.5 & 7.3 & 5.2 & 9.6 & 6.5 \\
\midrule
GPT-4o & \textcolor{gray!135}{23.1} & \textcolor{gray!135}{15.4} & \textcolor{gray!135}{8.3} & \textcolor{gray!135}{17.5} & \textcolor{gray!135}{6.7} & \textcolor{gray!135}{6.0} & \textcolor{gray!135}{4.2} & \textcolor{gray!135}{5.5} & \textcolor{gray!135}{2.7} & \textcolor{gray!135}{1.2} & \textcolor{gray!135}{3.2} & \textcolor{gray!135}{2.6} & \textcolor{gray!135}{39.0} \\
DeepSeek-R1-671B & \textcolor{gray!135}{40.5} & \textcolor{gray!135}{21.2} & \textcolor{gray!135}{5.2} & \textcolor{gray!135}{25.2} & \textcolor{gray!135}{5.0} & \textcolor{gray!135}{11.8} & \textcolor{gray!135}{11.3} & \textcolor{gray!135}{10.0} & \textcolor{gray!135}{8.5} & \textcolor{gray!135}{8.1} & \textcolor{gray!135}{9.3} & \textcolor{gray!135}{8.6} & \textcolor{gray!135}{42.4} \\
o1-preview$^\dagger$ & \textcolor{gray!135}{-} & \textcolor{gray!135}{-} & \textcolor{gray!135}{-} & \textcolor{gray!135}{-} & \textcolor{gray!135}{11.9} & \textcolor{gray!135}{10.4} & \textcolor{gray!135}{7.9} & \textcolor{gray!135}{9.9} & \textcolor{gray!135}{12.9} & \textcolor{gray!135}{8.1} & \textcolor{gray!135}{6.6} & \textcolor{gray!135}{11.1} & \textcolor{gray!135}{42.7} \\
\midrule
\multicolumn{14}{l}{\textit{\textbf{Single-Capability Enhanced}}} \\
Vanilla RAG & 40.5 & 21.2 & 5.2 & 25.2 & 57.4 & 44.6 & 40.0 & 46.0 & 10.6 & 3.7 & 11.6 & 9.6 & 72.5 \\
Search-o1 & 45.3 & 25.8 & 5.3 & 29.1 & \textbf{70.2} & 44.6 & 40.0 & 49.0 & 13.0 & 8.5 & 12.6 & 12.2 & 74.0 \\
WebThinker & \underline{50.0} & \underline{34.9} & \underline{10.5} & \underline{36.2} & 55.3 & \underline{53.0} & \underline{50.0} & \underline{52.5} & 13.9 & 9.7 & 12.6 & 13.0 & \underline{78.0} \\
CodeAct & 26.2 & 15.1 & 0.0 & 16.5 & 6.4 & 4.8 & 4.3 & 5.0 & 9.9 & 6.1 & 10.5 & 9.4 & 7.5 \\
Multimodal Enhanced & 23.8 & 9.1 & 0.0 & 12.6 & 4.3 & 0 & 4.3 & 4.0 & 9.3 & \underline{9.8} & 8.4 & 9.2 & 10.5 \\
\midrule
\multicolumn{14}{l}{\textit{\textbf{Multi-Capability Enhanced}}} \\
Plan-and-Solve & 28.6 & 18.2 & 0.0 & 18.9 & 44.7 & 33.7 & 24.3 & 33.0 & 10.2 & 4.9 & 7.3 & 8.8 & 57.5 \\
ReAct & 45.3 & 28.8 & 5.2 & 30.7 & 46.8 & 31.3 & 31.4 & 35.0 & 12.7 & \textbf{11.0} & \textbf{20.0} & \underline{13.8} & 73.5 \\
\rowcolor[RGB]{236,244,252} 
HiRA (ours) & \textbf{61.9} & \textbf{37.9} & \textbf{15.8} & \textbf{42.5} & \underline{59.6} & \textbf{54.2} & \textbf{51.4} & \textbf{54.5} & \textbf{15.2} & \textbf{11.0} & \underline{13.7} & \textbf{14.2} & \textbf{81.5} \\
\bottomrule
\end{tabular}
\caption{Overall performance on various deep search tasks, with accuracy results for each dataset obtained using llm-as-judge. For 32B models, the best results are indicated in \textbf{bold}, and the second-best results are \underline{underlined}. Results from larger or closed-source models are presented in \textcolor{gray!135}{gray} for reference. For the GAIA dataset, queries without files are used to ensure a fair comparison with the baseline.}
\label{tab:main_result}
\end{table*}

\paragraph{$\bullet$ Domain-Specialized Executors}
To cover the diverse capabilities required for deep search tasks, we design three orthogonal agent capability dimensions, ensuring the system can handle complex and varied deep search scenarios:
\begin{itemize}
\item \textbf{Information acquisition:} This dimension is responsible for acquiring and integrating information from the web.
\item \textbf{Cross-Modal understanding:} This dimension handles the understanding and fusion of multimodal information, capable of processing data from different modalities such as images, videos, and audio.
\item \textbf{Computational reasoning:} This dimension handles mathematical computation, file processing, and other computational reasoning tasks, capable of transforming abstract problems into executable code solutions.
\end{itemize}

Based on these three dimensions, we implement four reasoning model-driven, specialized agents. For \textbf{information acquisition}, we design two search agents with different exploration depths: one based on a simple RAG pipeline that performs single retrieval for subtasks followed by reasoning, and another based on the WebThinker implementation~\cite{Li2025WebThinker} that can perform deep search and information acquisition on the internet. The combination of these two approaches enables flexible solutions for both simple and complex tasks. For \textbf{cross-modal understanding}, we embed multimodal models as tools within the reasoning model's inference process to achieve dynamic understanding of information in multimodal data. For \textbf{computational reasoning}, we embed code interpreters into the reasoning process.

For the aforementioned reasoning-driven agents, their reasoning process follows the tool-augmented reasoning execution flow, capable of dynamically outputting special tokens during reasoning to trigger corresponding tools. The reasoning process is:
\begin{equation*}
	P(\mathcal{O}^{(k)} \mid s_k, \mathcal{T}, \mathcal{M}_k) = \sum_{t=1}^{T_k}P(\mathcal{O}_t^{(k)} \mid \mathcal{O}_{<t}^{(k)}, \{\mathcal{T}_j\}_{<t}, \cdot),
\end{equation*}
where $\mathcal{O}^{(k)}$ represents the reasoning process of expert agent for subtask $s_k$, $\mathcal{M}_k$ represents the memory related to $s_k$, $\mathcal{T}$ represents the tools used in the process (e.g., code interpreter), and $\{\mathcal{T}_j\}_{<t}$ represents all tool invocation results before time step $t$. Based on this process, we can embed arbitrary tools into the reasoning model's inference process. More details can be found in the appendix.

\subsection{Inference Process of \ours{}}
The inference process of \ours{} follows an agentic reasoning approach. For a given question, the inference process begins with reasoning by the meta agent. During reasoning, the meta planner decodes special token pairs, wrapping the subtasks to be executed between them. The coordinator then processes and distributes the subtasks to expert agents for execution. After the expert agents perform reasoning and multiple rounds of tool invocation, subtask execution results are obtained.

This reasoning process and results are then processed by the coordinator through the reasoning distillation process to obtain refined results, which are integrated into the meta planner's reasoning chain to continue generation. During this process, the meta planner dynamically adjusts its plan and corresponding subtasks to be distributed based on execution results, until all information has been collected and the final answer is provided. The detailed example can be found in the appendix.

\section{Experimental Settings}

\paragraph{Tasks and Datasets}

To thoroughly assess our method on deep search tasks, we expand prior setups by introducing multimodal and file-based scenarios. We evaluate on:  (1) \textbf{GAIA}~\cite{GAIA}, covering multi-step reasoning and retrieval, using all validation samples across text-only, multimodal, and file-based categories;  (2) \textbf{WebWalkerQA}~\cite{2501_WebWalker}, testing web navigation and extraction in English and Chinese, with 200 sampled questions;  (3) \textbf{SimpleQA}~\cite{simpleqa}, evaluating factual and broad knowledge, with 200 sampled questions; and  (4) \textbf{Humanity's Last Exam}~\cite{HLE}, a high-difficulty benchmark requiring complex reasoning and external retrieval, with 500 validation samples.  All evaluations use Qwen2.5-72B-Instruct~\cite{qwen2.5} as LLM judge to compute accuracy. Complete details are included in the appendix.

\paragraph{Baselines}
We compare our method with three baseline categories:
(1) Direct reasoning: Using models’ native reasoning abilities, including both open-source models (Qwen3-32B~\cite{qwen3}, QwQ-32B~\cite{qwen_qwq}, DeepSeek-R1-32B~\cite{deepseek-r1}) and commercial ones (GPT-4o~\cite{gpt_4o_system_card}, DeepSeek-R1~\cite{deepseek-r1}, o1-preview~\cite{2409_openai_o1}).
(2) Single-capability enhanced: Augmenting reasoning with one specialized tool—e.g., search (Search-o1~\cite{searcho1}, WebThinker~\cite{Li2025WebThinker}), code execution (CodeAct~\cite{wang2024codeact}), or multimodal reasoning.
(3) Multi-capability reasoning: Integrating multiple tools or structured workflows, including Plan-and-Solve~\cite{wang2023planandsolvepromptingimprovingzeroshot} and ReAct~\cite{react}. These use the same toolset as ours for fair comparison. More details are included in appendix.

\section{Experimental Results}

\begin{table}[t]
\centering
\small
\setlength\tabcolsep{1.2pt} 
\begin{tabular}{lcccccc}
\toprule
Method & GAIA-B & GAIA-F & Web. & HLE & Simp. &  Avg. \\
\midrule
\rowcolor[RGB]{236,244,252} 
\textbf{\ours{}} & \textbf{42.5} & \textbf{42.1} &  \textbf{54.5} & \textbf{14.2} & \textbf{81.5} & \textbf{44.9} \\
\midrule
\multicolumn{7}{l}{\textit{\textbf{Coordinator Layer}}} \\
~~w/o Reasoning Trans. & 33.9  & 36.8 & 44.5 & 10.4 & 76.5 & 40.4 \\
~~w/o Memory & 37.8 & 31.6 & 52.0 & 11.8 & 79.0 & 42.4  \\
\multicolumn{7}{l}{\textit{\textbf{Executor Layer}}} \\
~~w/o Search & 15.7 & 31.6 &  4.0 & 12.4 & 9.5 & 14.6\\
~~w/o Code & 33.9 & 28.9 & 51.5 & 12.8 & 76.5 & 40.7 \\
~~w/o Multimodal & 36.2 & 36.8  & 55.0 & 13.6 & 81.0 & 44.5 \\
\bottomrule
\end{tabular}
\caption{Ablation studies of \ours{}, showing results for each layer. GAIA-B refers to GAIA queries without associated files, while GAIA-F refers to the subset with files.}
\label{tab:ablation_study}
\end{table}

\subsection{Main Results}
As shown in Table~\ref{tab:main_result}, we evaluate our method on the deep search tasks against several baselines that integrate reasoning with tool usage. We have the following observations:
(1) \textbf{Overall performance superiority}: Our method consistently outperforms all baseline methods. It significantly surpasses direct reasoning models without tool usage and achieves notable improvements over existing tool-augmented approaches. Compared to the strongest search agent baseline WebThinker, our approach demonstrates substantial advantages on both complex tasks (GAIA and HLE).
(2) \textbf{Hierarchical reasoning design advantages}: Results show that our hierarchical design achieves better performance when using the same set of tools. For Plan-and-Solve which relies on a fixed plan executed sequentially, performs poorly (e.g., 18.9 in GAIA), highlighting the necessity of dynamic planning during reasoning. While ReAct enables dynamic planning by integrating multiple tools into the reasoning chain, its performance suffers in multi-tool scenarios such as GAIA due to the overhead of tool selection and noisy intermediate tool outputs. In contrast, our method outperforms WebThinker on GAIA, which requires diverse capabilities, while also achieving superior results on general web search tasks.
(3) \textbf{Robustness across task complexity}: Our framework shows moderate gains on simpler tasks (e.g., SimpleQA and WebWalkerQA), but exhibits much larger improvements on more complex tasks, demonstrating its strength in handling complex reasoning scenarios.

\begin{figure}[!t]
\centering
\includegraphics[width=0.95\linewidth]{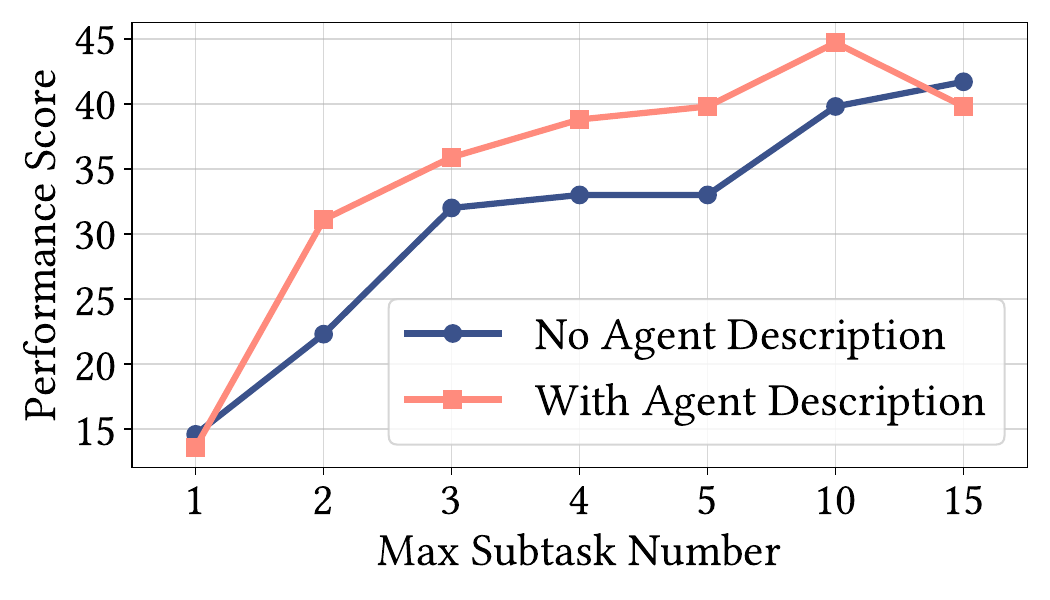}
\caption{Performance comparison on whether the expert agent description is provided to the meta planner and maximum number of sub-tasks limit.}
\label{fig:agent_description}
\end{figure}

\begin{figure}[!t]
\centering
\includegraphics[width=0.95\linewidth]{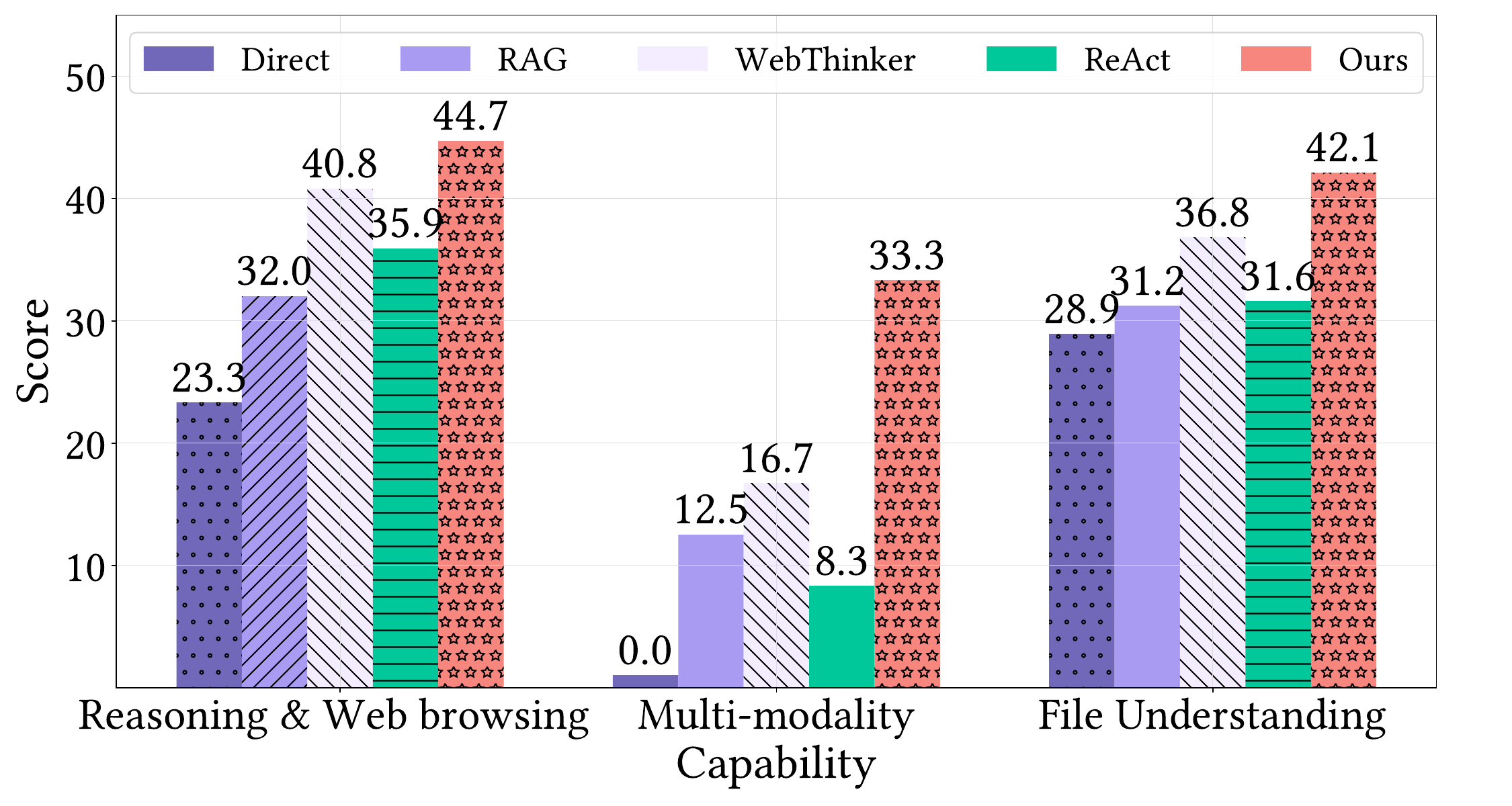}
\caption{Comparison of our method with the baseline on three GAIA subsets, evaluating performance across different dimensions of capability.}
\label{fig:capability_analysis}
\end{figure}

\subsection{Ablation Study}
We conduct ablation studies to investigate the contribution of each component within the framework by removing modules from both the Coordinator Layer and the Executor Layer. Results are shown in Table~\ref{tab:ablation_study}.
In the Coordinator Layer, removing the Reasoning Transfer mechanism leads to significant performance drops, especially on complex reasoning tasks, with nearly 30\% decrease in performance. In comparison, the Memory mechanism has a smaller impact, except on file-related tasks (approx. 10\% drop), indicating that the resource component in memory effectively supports information propagation.

In Executor Layer, removing individual expert agents leads to substantial performance degradation. Among them, removing the Search Agent causes the most severe drop across all datasets, highlighting the essential role of information seeking on web. Similarly, removing the Code Agent significantly impacts performance on multi-functional datasets such as GAIA, showing its importance for general-purpose tasks. Removing the Multimodal Agent results in a slight drop on GAIA, where some cases require multimodal capabilities.

\subsection{Generalization and Effectiveness of Meta Planner}
In our framework, the meta-planner receives expert capability information for better subtask planning, while also limiting the number of subtasks to control computational cost. These two factors affect both generalization (supporting more agents) and effectiveness (achieve higher performance). To evaluate both aspects, we introduce an additional setting where no expert information is provided to the meta-planner, and we vary the maximum number of allowed subtasks. Results are shown in Figure~\ref{fig:agent_description}, leading to two key observations:
(1) \textbf{Decoupled meta-planner design}: Even without expert information, the meta-planner achieves comparable performance. This suggests that, within our architecture, the meta-planner and expert agents are relatively decoupled—the generated subtasks primarily depend on the task semantics rather than the specific agents, enabling better scalability to new agents.
(2) \textbf{Subtask scaling trade-off}: As the number of subtasks increases, performance first improves then degrades. This resembles an inference-time scaling effect: allowing more subtasks enables deeper reasoning chains, while overly restricting subtasks may prevent sufficient exploration. However, excessive subtasks may introduce inefficient plans, eventually hurting performance.

\begin{figure}[!t]
\centering
\includegraphics[width=0.99\linewidth]{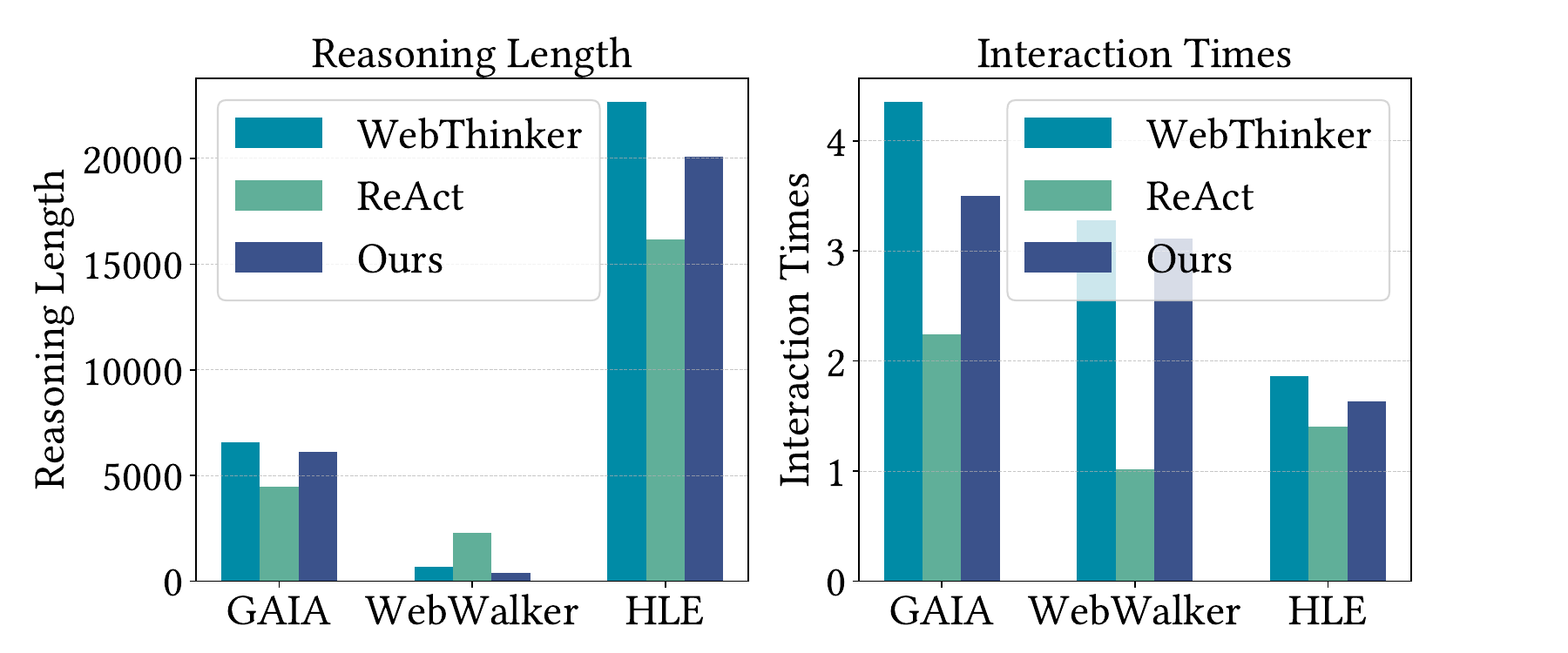}
\caption{Comparison of different methods in terms of reasoning length (number of output tokens during model inference) and interaction times (number of interactions with the environment during inference) in three datasets.}
\label{fig:reasoning_efficiency}
\end{figure}

\subsection{Capability Analysis}
Beyond the main experiments, we further evaluate our system on additional subsets of GAIA to assess its capabilities beyond web search (e.g., multimodal understanding, file understanding), which are also critical for more comprehensive information acquisition. As shown in Figure~\ref{fig:capability_analysis}, our method achieves the best performance across all capability dimensions, demonstrating its broad applicability.
While baselines like ReAct perform well in purely textual reasoning and web browsing, our integration of the DeepSearch agent yields superior results. Moreover, due to our multimodal architecture, we achieve additional gains in non-textual tasks. It is worth noting that, although ReAct can invoke multimodal and code tools, its planning model struggles to coordinate multiple tools simultaneously, often leading to sub-optimal performance (e.g., 8.3 vs. 12.5 in multimodal tasks), in some cases even underperforming pure search-based approaches.

\subsection{Efficiency Analysis}
To assess the overall efficiency of our method, we compared the number of inference tokens and the number of interactions with the environment between \ours{} and baselines across three datasets of varying difficulty. Based on the result in Figure~\ref{fig:reasoning_efficiency} and Table~\ref{tab:main_result}, we make two observations:
(1) Overall, more difficult datasets tend to result in longer reasoning chains, but they do not necessarily lead to more interactions with the environment, which may also be influenced by the nature of the task itself.
(2) Compared to WebThinker, which directly integrates the search function into the main reasoning chain, our hierarchical reasoning structure results in shorter reasoning chains and fewer interaction times, indicating that our approach is more efficient in each subtask call than directly invoking tools.
(3) The ReAct method, which integrates multiple functionalities into a single model, leads to fewer tool calls and insufficient reasoning, which may be due to interference between the descriptions of multiple tools, resulting in suboptimal performance.

\section{Conclusion}
In this paper, we propose \ours{}, a hierarchical reasoning framework that overcomes limitations of monolithic models in deep search tasks. Our multi-agent architecture combines a Meta reasoning planner, Adaptive Reasoning Coordinator, and Domain-Specialized Executors to enable scalable, modular reasoning through specialized cognitive components rather than rigid pipelines. Using dual-channel memory and coordinated task routing, \ours{} supports coherent knowledge synthesis and dynamic integration of diverse tool capabilities. Experiments across five complex, multi-modal deep search tasks show \ours{} significantly outperforms conventional RAG systems and existing agent-based methods, highlighting the effectiveness of our decoupled design.

\bibliography{reference}

\newpage
\clearpage
\clearpage
\appendix
\onecolumn

\section*{Appendix}
\subsection{Datasets}
(1) \textbf{GAIA}: A benchmark for evaluating general-purpose AI assistants, where questions require multi-step reasoning and external information retrieval. We use all samples from the validation set and categorize them into three types based on input and required information: text-only, multimodal, and with-file. 
Specifically, with-file represents queries where the input question comes with accompanying files, text-only represents queries that only require text capabilities, which are filtered through annotated metadata, and the remaining queries are categorized into multimodal.
(2) \textbf{WebWalkerQA}: A benchmark designed to evaluate models' ability to navigate web pages and extract information. Questions require aggregating knowledge from single or multiple web pages and cover both English and Chinese queries. We sample 200 questions from the test set.
(3) \textbf{SimpleQA}: A benchmark testing models' factual accuracy, examining knowledge breadth and factual capabilities. Since models inherently suffer from hallucination issues and may struggle on this dataset, this benchmark effectively tests models' ability to answer questions by incorporating external knowledge. We sample 200 questions from the test set.
(4) \textbf{Humanity's Last Exam}: A high-difficulty dataset requiring complex reasoning, containing academic problems across multiple domains including mathematics, physics, and computer science. Models need to retrieve relevant external information to support their reasoning. We sample 500 questions from the validation set. To simplify the calculation of scores, we merged the original subsets of the HLE dataset into three major categories: (1) Natural Science, which includes the original Math, Bio, Physics, and Chemistry; (2) Computational Science and Engineering, covering CS and Engineering; and (3) Social Fields, including Humanities and Other.

\subsection{Evaluation Metrics}
To fairly evaluate the effectiveness of our methods, we adopted the LLM-as-Judge approach across all datasets, using Qwen2.5-72B-Instruct to assess the consistency between model-generated answers and golden answers. The instruction used here follows Webthinker~\cite{Li2025WebThinker}, as shown in following instruction~\ref{eval_normal}. It should be noted that in the construction process of the WebWalker dataset we used, each question's answer is derived solely from the corresponding annotated webpage, without collection and organization from the web, which leads to biased and inconsistent golden answers. This results in models providing more detailed answers being incorrectly judged as wrong (due to inconsistency with standard answers). Therefore, in evaluation of WebWalker, we optimized the instruction template, as shown in following instruction~\ref{eval_webwalker}.

\begin{tcolorbox}[colback=myred!5!white,colframe=myred!75!black,title=Evaluation Instruction for Normal QA Tasks,enhanced jigsaw,breakable,label=eval_normal]
\begin{VerbatimWrap}
You are an evaluation assistant. Please determine if the model output is equivalent to the labeled answer.

Question: {question}

Labeled Answer: {labeled_answer}

Model Output: {pred_answer}

Did the model give an answer equivalent to the labeled answer? Please respond with "Correct" if they are equivalent, or "Incorrect" if they are not equivalent. Do not include any other text.
\end{VerbatimWrap}
\end{tcolorbox}

\begin{tcolorbox}[colback=myred!5!white,colframe=myred!75!black,title=Evaluation Instruction for WebWalker,enhanced jigsaw,breakable,label=eval_webwalker]
\begin{VerbatimWrap}
You are an evaluation assistant. Please determine if the predicted answer is equivalent to the labeled answer in terms of core content.

Question: {question}

Labeled Answer: {labeled_answer}

Predicted Answer: {pred_answer}

Evaluation Guidelines:
1. Mark as "Correct" if the predicted answer contains all key information from the labeled answer, even if it provides additional details
2. Mark as "Correct" if the predicted answer matches the core meaning of the labeled answer, regardless of different phrasing
3. Only mark as "Incorrect" if the predicted answer misses critical information or contains incorrect information

Please respond with only "Correct" or "Incorrect". Do not include any other text.
\end{VerbatimWrap}
\end{tcolorbox}

\subsection{Implementation Details}
\subsubsection{Experiment Details}
Following current work, we use QwQ-32B as the base model for both the meta reasoning planner and expert agents. We employ a same-sized Qwen2.5-Instruct model as the Adaptive Reasoning Coordinator. For all models, we set temperature to 0.7, top\_p to 0.95, top\_k to 20, and the model context window to 128k tokens.
Our search tool uses the Bing Web Search API with US-EN as the region setting and retrieves the top-10 search results. We employ Qwen2.5-Omni-7B~\cite{xu2025qwen25omnitechnicalreport} as our multimodal tool, powered by Qwen's official API. We construct a sandbox based on a Python interpreter as our code execution environment, with necessary security restrictions to ensure safe operation. In the main experiments, we set the maximum number of subtasks for the meta reasoning planner to 10.

\subsubsection{Baseline Details}
All our baselines are implemented based on QwQ-32B as the base model. The implementation details for each baseline are described below:
\begin{itemize}
    \item \textbf{Vanilla RAG}: We directly utilize the original question for search and then provide the retrieved results to the generator for answer generation.
    \item \textbf{Search-o1}: This approach integrates search capabilities during the reasoning process, dynamically generating queries and performing searches. Our implementation is based on the official Search-o1 repository.
    \item \textbf{WebThinker}: This method enables dynamic searching, link clicking, and report generation during the reasoning process. Since our evaluation does not involve report generation, we only utilize the question-answering functionality. Our implementation is likewise based on the official repository.
    \item \textbf{CodeAct}: This approach can invoke code tools to assist reasoning during the inference process. We employ a Python Sandbox as the code tool and design custom prompts for implementation, with the instruction shown in following instruction~\ref{instruction_code}.
    \item \textbf{Multimodal Enhanced}: Similar to CodeAct, we replace the Python Sandbox with the multimodal tool used in our method (i.e., the Qwen2.5-omni-7B model) as a tool that can be dynamically invoked during reasoning, with the instruction shown in following instruction~\ref{instruction_mm}.
    \item \textbf{Plan-and-Solve}: This method first comprehends the original question and generates an overall plan, then executes this plan step-by-step until reaching the final result. The instruction for plan generation is shown in following instruction~\ref{plan_and_act}. During plan execution, to ensure fair comparison with our method, we automatically determine the most appropriate expert agent for each step in the plan through the model, using the same set of expert agents as in our approach.
    \item \textbf{ReAct}: This approach allows the model to directly utilize multiple tools as needed during reasoning to assist inference. Tool descriptions are provided to the model at the beginning, and tool outputs are directly inserted into the chain of thought.
\end{itemize}

\begin{tcolorbox}[colback=myred!5!white,colframe=myred!75!black,title=Instruction Template for Query Planning in Simple Search Agent,enhanced jigsaw,breakable,label=query_planning]
\begin{VerbatimWrap}
You are a reasoning assistant. Your task is to generate a detailed query plan for answering the user's question by breaking it down into sub-queries. Each query should be a precise and suitable query for web search.

Please analyze the question and break it down into multiple sub-queries that will help gather all the necessary information to answer it completely. 

Remember:
1. Each sub-query should be a precise and suitable query for web search.
2. Subqueries should cover different aspects to avoid duplicate searches.
3. Each subquery needs to be necessary, do not add irrelevant ones. You need to use as few subqueries as possible to obtain accurate results.


Output your query plan in JSON format as follows:

```json
{{
    "query_plan": [
        "sub-query-1",
        "sub-query-2",
        ...
    ]
}}
```

Task: {question}
\end{VerbatimWrap}
\end{tcolorbox}

\begin{tcolorbox}[colback=myred!5!white,colframe=myred!75!black,title=Instruction Template for Answer Generation in Simple Search Agent,enhanced jigsaw,label=instruction_simple_search,breakable]
\begin{VerbatimWrap}
You are a knowledgeable assistant that uses the provided documents and previous exploration results as your current memory to answer the user's question step by step.

You need to carefully analyze the given context step by step and then provide a reliable, detailed answer to the task. If you find that there is no valid information in the document, please directly state that there is no valid information.

Your answer needs to include specific analysis and relevant citations (detail page url and title).

Question: {question}
Documents: {documents}
\end{VerbatimWrap}
\end{tcolorbox}

\begin{tcolorbox}[colback=myred!5!white,colframe=myred!75!black,title=Instruction Template for Answer Generation in Computation Reasoning Agent,enhanced jigsaw,label=instruction_code,breakable]
\begin{VerbatimWrap}
You are a reasoning assistant with the ability to execute Python code to help you answer the user's question accurately. 
You have special tools: 
- To execute a python code: write <|begin_code_call|>python your python code here <|end_code_call|>. Then, the Python code interpreter will execute the python code, and provide you with execution result in the format <|begin_code_call_result|> ...code call results... <|end_code_call_result|>. If the code encounters errors while running, try to fix them and give another code call. You can repeat the code execution process multiple times if necessary. The maximum number of code execution attempts is limited to {MAX_CODE_CALL_NUM}. Once you have all the information you need, continue your reasoning. You are given previous exploration results as your current memory, please first check the current memory for relevant information before making search. But you need to be aware that the facts involved in memory may not be comprehensive. 

Example: {example}

Remember: 
- Use <|begin_code_call|> to start Python code and end with <|end_code_call|>. 
- Always explain your reasoning before and after executing code. 
- When done with code execution, provide a clear final answer or solution. Please answer the following question step by step. When encountering scenarios that require computation, try to use code for verification instead of solely using own knowledge. You should provide your final answer in the format \boxed{{YOUR_ANSWER}}. 

Current Memory: {current_memory} 
Question: {task_info}
\end{VerbatimWrap}
\end{tcolorbox}

\begin{tcolorbox}[colback=myred!5!white,colframe=myred!75!black,title=Instruction Template for Answer Generation in Multimodal Reasoning Agent,enhanced jigsaw,label=instruction_mm,breakable]
\begin{VerbatimWrap}
You are an AI assistant with multimodal understanding capabilities. You can analyze images, video and audio to answer user questions. 
You have access to a special multimodal tool: - To analyze an image/video/audio and answer questions about it, use the format: <|begin_multimodal_call|> data: [path of image/video/audio] question: [your specific question] <|end_multimodal_call|> The system will provide analysis results in the format: <|begin_multimodal_result|> ...analysis results... <|end_multimodal_result|> You can ask multiple questions about different aspects of the image/video/audio if needed. The maximum number of multimodal analysis calls is limited to {MAX_MM_CALL_NUM}. 
You are given previous analysis results as your current memory. Please check the current memory for relevant information before making new analysis requests. Example: {example}

Remember: - Always explain your reasoning before and after multimodal analysis - Provide your final answer in \boxed{YOUR_ANSWER} format 
Current Memory: {current_memory} 
Question: {task_info}
\end{VerbatimWrap}
\end{tcolorbox}

\begin{tcolorbox}[colback=myred!5!white,colframe=myred!75!black,title=Instruction Template for Plan-and-Act,enhanced jigsaw,breakable,label=plan_and_act]
\begin{VerbatimWrap}
You are a reasoning assistant. For the given task, come up with a simple step by step plan. 
This plan should involve individual tasks, that if executed correctly will yield the correct answer. Do not add any superfluous steps. 
The result of the final step should be the final answer. Make sure that each step has all the information needed - do not skip steps.

Give your plan in JSON format as follows:
```json
{{
    "plan": [
        "step1", # each step is a string of the task description
        "step2",
        ...
    ]
}}

Remember:
1. Each step should be a self-contained task, describe the task in detail.
2. The result of the final step should be the final answer.
3. Output your plan in JSON format.
```

Task:
{question}
\end{VerbatimWrap}
\end{tcolorbox}

\subsubsection{Details of Domain-Specific Agents}
In our experiments, we design four types of expert agents to implement different functionalities and accomplish various types of tasks. The specific design details for all agents are as follows:
\begin{itemize}
    \item \textbf{Simple Search Agent}: Responsible for searching and integrating information on the internet, suitable for simple and efficient information gathering and fact verification. Its workflow includes understanding the given subtask and generating a query plan, then conducting searches and collecting information to provide answers. The instruction templates for query planning and answer generation are shown in following instruction~\ref{query_planning} and following instruction~\ref{instruction_simple_search}, respectively.
    \item \textbf{Deep Search Agent}: Responsible for conducting deep exploration and information collection on the internet, requiring capabilities such as utilizing search engines and clicking webpage links. We directly adopt the implementation of WebThinker~\cite{Li2025WebThinker} as our deep search agent.
    \item \textbf{Computational Reasoning Agent}: This agent can autonomously invoke a Python interpreter during the reasoning process to assist its thinking and information collection. The covered functionalities include but are not limited to: reading files, downloading files, and obtaining webpage information through code. We follow the approach of Search-o1~\cite{searcho1}, using the QwQ-32B reasoning model to drive a Python tool encapsulated in a sandbox, with the specific instruction shown in following instruction~\ref{instruction_code}.
    \item \textbf{Multimodal Agent}: Possesses the capability to invoke multimodal tools during the reasoning process. To simplify deployment and reduce usage complexity, we employ Qwen-omni-7B as our multimodal tool, which can understand files of various modalities. We embed it into the model's reasoning process, enabling the model to invoke this tool during thinking to handle multimodal scenarios, with the specific instruction shown in following instruction~\ref{instruction_mm}.
\end{itemize}

\subsubsection{Details of Coordinator}
We employ Qwen2.5-32B-Instruct as the base model for our coordinator, which primarily incorporates three functions: 
(1) \textbf{Reasoning Transfer Process}: The model is assigned a subtask and subsequently identifies the most suitable expert agent through classification. During this process, comprehensive information about all available expert agents is provided to the model to facilitate its understanding of each agent's capabilities. The specific instruction is formulated as follows:

\begin{tcolorbox}[colback=myred!5!white,colframe=myred!75!black,title=Instruction for Reasoning Transfer Process,enhanced jigsaw,label=reasoning_transfer,breakable]
\begin{VerbatimWrap}
You are an agent selection system. Analyze the given task and select the most suitable agent based on:

1. Required Capabilities:
- What specific skills/knowledge does this task demand?
- How well does each agent's expertise match these requirements?

2. Task Difficulty:
- Complexity level (simple fact vs multi-step problem-solving). You should consider the effective time cost of each agent.
- Depth of analysis needed (surface information vs deep exploration)
- You need to choose the model **that can complete the task** with the lowest cost/complexity as much as possible.

**Only output the JSON format** with the following fields:
- reason: The reason for selecting the agent
- selected_agent_name: The name of the selected agent

Example Output: {example}
Agents Available: {agent_info}

Task: {task}
Analyze the task and respond **ONLY the json format**, without any additional explanation.
\end{VerbatimWrap}
\end{tcolorbox}

(2) \textbf{Reasoning Distillation Process}: Upon completion of a subtask by an expert agent, the coordinator comprehends and summarizes the agent's reasoning trajectory, subsequently refining it into a distilled reasoning process and final conclusion for reporting to the meta-planner. This mechanism enables the meta-planner to verify the logical validity of the reasoning pathway and evaluate the derived results.

(3) \textbf{Memory Construction}: Beyond the conclusion, the coordinator extracts and records key findings from the expert agent's reasoning process into the global memory. Specifically, we design the coordinator to capture two memory components: facts and resources. To enhance computational efficiency, this functionality is implemented within the same instruction as the reasoning distillation process. The detailed instruction is formulated as follows:

\begin{tcolorbox}[colback=myred!5!white,colframe=myred!75!black,title=Instruction for Reasoning Distillation Process and Memory Construction,enhanced jigsaw,label=reasoning_transfer,breakable]
\begin{VerbatimWrap}
You are a specialized Summarization Agent. Your role is to analyze a problem and a model's thinking process, then produce a structured summary with two key components:

1. CONCLUSION: Create a concise string that captures:
   - reasoning_process: A concise string outlining the necessary reasoning steps in logical order, including key actions, searches, and findings.
   - final_conclusion: A concise string that captures the final answer and conclusion of the given task.

2. MEMORY: **Organize useful information for future tasks** (only include the information that highly likely to be useful for future tasks, don't include current memory facts):
   - fact_memory: List important facts discovered during reasoning
     * Each entry must include both content AND source
     * Sources must be specific (exact URLs, specific document names, or "Model Inference" for reasoned conclusions)
     * Consolidate related facts into single entries to avoid fragmentation. Only keep facts that are relevant to the future tasks.
     * If no valid source is found, set it as [Source: Not Specified]
   - resource_memory: Map useful resources as "description": "path" pairs
     * Paths must be complete, valid URLs or precise variable names
     * Descriptions should be clear and specific
     * Include only verified, accessible resources
     * If no valid resources exist, set resource_memory as an empty dictionary

Produce ONLY a properly formatted JSON object with these components. Include nothing else.

Example output: {example}

Reasoning Chain:
{reasoning_chain}

Task: {task_description}
\end{VerbatimWrap}
\end{tcolorbox}

(4) \textbf{Memory Utilization}: During subtask delegation, the coordinator retrieves relevant memory segments from the global memory and provides them to the expert agent alongside the subtask. The instruction is structured as follows:

\begin{tcolorbox}[colback=myred!5!white,colframe=myred!75!black,title=Instruction for Memory Utilization,enhanced jigsaw,label=reasoning_transfer,breakable]
\begin{VerbatimWrap}
You are an assistant specialized in filtering memory based on a specific task. Your task is to analyze the given memory and select ONLY the most task-relevant memories, with a strict maximum limit of 5 entries.

Key Requirements:
1. Relevance First:
   - Each selected memory MUST have a direct and strong connection to the current task
   - Reject memories that are only tangentially or weakly related
   - If there are fewer than 5 highly relevant memories, select only those that are truly relevant

2. Quality Control:
   - Filter out any memories with invalid or suspicious URLs
   - Remove memories about failed attempts or negative experiences
   - Exclude memories that contain speculative or unverified information

3. Output Format:
   - Output the filtered memories in the following format:
     ```
     Memory Fact 1: [memory1]
     Memory Fact 2: [memory2]
     ...
     ```
Remember: It's better to return fewer but highly relevant memories than to include marginally related ones just to reach 5 entries.

Memory:
{memory}

Task:
{task}

Filtered Memory:
\end{VerbatimWrap}
\end{tcolorbox}

\subsection{Time Efficiency Analysis}
Despite the multi-agent architecture, HiRA achieves time efficiency comparable to deep search baselines (e.g. WebThinker, ReAct). This comparability stems from two primary factors: (1) \textit{Lightweight agent utilization}: Our domain-specialized executors employ efficient naive RAG approaches for simpler subtasks, avoiding the computational complexity of deep web exploration when unnecessary. (2) \textit{Reduced tool interaction overhead}: The hierarchical delegation enables more targeted tool usage, with each expert agent making fewer but more precise tool calls compared to monolithic reasoning approaches.

\subsection{Case Study}
The following is a case from the GAIA dataset that demonstrates the capability of our method, shown in Table~\ref{tab:case}.

\begin{table*}[!t]
    \centering
    \caption{An example of our method inference process from the GAIA dataset.}
    \fontsize{9pt}{11pt}\selectfont
    \begin{tabular}{p{0.98\linewidth}}
    \midrule
        \rowcolor{gray!20}\textbf{Example \# from GAIA Dataset} \\
    \midrule
        \textbf{Question:} \\
        In terms of geographical distance between capital cities, which 2 countries are the furthest from each other within the ASEAN bloc according to wikipedia? Answer using a comma separated list, ordering the countries by alphabetical order. \\
        \textbf{Labeled Answer:} \\
        Indonesia, Myanmar \\
    \midrule
        \rowcolor{gray!20}
        \textbf{Inference Process}\\
    \midrule
        Okay, so the user wants to know which two ASEAN countries have their capital cities the farthest apart geographically. Hmm, first I need to recall what countries are part of ASEAN. ... 
        The first subtask call would be: \\
       \textbf{\textless begin\_call\_subtask\textgreater}
       Retrieve the list of ASEAN member countries along with their respective capital cities from Wikipedia. 
      \textbf{\textless end\_call\_subtask\textgreater}
        \\

        \textbf{\textless begin\_subtask\_result\textgreater} 
        Result from Search Agent:...The list of ASEAN member countries and their respective capital cities is: 1. Brunei - Bandar Seri Begawan, 2. Cambodia - Phnom Penh, 3. Indonesia - Jakarta, 4. Laos - Vientiane, 5. Malaysia - Kuala Lumpur, 6. Myanmar - Nay Pyi Taw, 7. Philippines - Manila, 8. Singapore - Singapore, 9. Thailand - Bangkok, 10. Vietnam - Hanoi. 
        \textbf{\textless end\_subtask\_result\textgreater} \\

        Okay, got the list. Now, the next step is to get the coordinates for each capital. Since there are 10 capitals, doing a separate search for each would take 10 subtasks, which is too many. So better to use code to batch process. ...Therefore, the second subtask is: \\

\textbf{\textless begin\_call\_subtask\textgreater}
Write Python code to retrieve the geographic coordinates of each ASEAN capital city from Wikipedia using the Wikipedia API, calculate the pairwise distances between all capital cities using the Haversine formula, identify the pair with the maximum distance, and return the two countries in alphabetical order. 
\textbf{\textless end\_call\_subtask\textgreater}\\

\textbf{\textless begin\_subtask\_result\textgreater} 
Result from Code-Agent:Identified the list of ASEAN countries and their capitals. Wrote a function to fetch geographic coordinates for each capital using the Wikipedia API. Implemented the Haversine formula to compute pairwise distances between all capital cities. Tracked the pair of capitals with the largest distance. Sorted the two countries alphabetically and returned them.Final Conclusion:The two ASEAN countries with the farthest capital cities, in alphabetical order, are Indonesia and Vietnam. 
\textbf{\textless end\_subtask\_result\textgreater} \\

Wait, but is that correct? Let me think. Jakarta (Indonesia) and Hanoi (Vietnam). Are they the farthest apart?Alternatively, maybe Brunei and Indonesia's capitals are further apart? Let me think of approximate coordinates:Bandar Seri Begawan: approx 4.9°N, 115°EJakarta: ~-6.2°S, 106.8°EDistance between these would be roughly from around 4.9N to -6.2S, so total latitude difference of ~11.1 degrees.... Alternatively, since I can only use up to 10, and already used 2, perhaps proceed as follows:- Acknowledge that the first code result was incorrect due to the capital name, so need to adjust.- Re-run the code with corrected capital names, but to do that, I need to specify the correction in the code. Thus, the second code subtask would be: \\

\textbf{\textless begin\_call\_subtask\textgreater}
Write Python code to calculate the geographical distance between all pairs of ASEAN capital cities using their precise coordinates obtained from Wikipedia, correcting the capital of Myanmar to 'Naypyidaw'. The countries and capitals are: Brunei (Bandar Seri Begawan), Cambodia (Phnom Penh), Indonesia (Jakarta), Laos (Vientiane), Malaysia (Kuala Lumpur), Myanmar (Naypyidaw), Philippines (Manila), Singapore (Singapore), Thailand (Bangkok), Vietnam (Hanoi). Use the Haversine formula and return the two countries with the farthest capitals in alphabetical order.
\textbf{\textless end\_call\_subtask\textgreater}\\

\textbf{\textless begin\_subtask\_result\textgreater} 
Result from Code-Agent:Corrected the capital of Myanmar to Naypyidaw. ...\\The two countries with the farthest capitals in alphabetical order are ('Indonesia', 'Myanmar'). 
\textbf{\textless end\_subtask\_result\textgreater} \\

Ah, that would be the correct result. Thus, the second code subtask, after correcting the capital name, gives Indonesia and Myanmar.But since I already used two subtasks (the first search and the first code), and this is the third, that's okay.Wait, but according to the initial plan, the first code subtask might have failed because of the capital name, so I need to fix that....\\
But since the first code result was wrong, perhaps I need to adjust.So following this adjusted plan:First subtask (already done) gives the initial list, but with a typo in Myanmar's capital.Second subtask (search for correct capital name of Myanmar): \\

\textbf{\textless begin\_call\_subtask\textgreater}
Confirm the correct spelling of the capital city of Myanmar according to Wikipedia. 
\textbf{\textless end\_call\_subtask\textgreater}\\
\textbf{\textless begin\_subtask\_result\textgreater} 
Result from Search Agent:Analyzed multiple documents to determine the correct spelling of Myanmar's capital city. Document 1 from Simple English Wikipedia primarily uses 'Naypyidaw'. Document 2 from Britannica uses 'Nay Pyi Taw'. Document 3 from Britannica also uses 'Nay Pyi Taw' as the primary spelling. Document 4 from Wikiwand uses 'Naypyidaw' as the primary spelling and mentions 'Nay Pyi Taw' as the officially romanized version. Document 5 from Britannica provides context on naming conventions in Myanmar but does not directly address the capital city's spelling.Final Conclusion:The correct spelling of the capital city of Myanmar according to Wikipedia is 'Naypyidaw'. 
\textbf{\textless end\_subtask\_result\textgreater} \\

Okay, so the correct spelling is Naypyidaw.Now, use the code-agent with the corrected capital names.The code would now have the correct list:capitals = [('Brunei', 'Bandar Seri Begawan'),    ('Cambodia', 'Phnom Penh'),    ('Indonesia', 'Jakarta'),    ('Laos', 'Vientiane'),    ('Malaysia', 'Kuala Lumpur'),    ('Myanmar', 'Naypyidaw'),    ('Philippines', 'Manila'),    ('Singapore', 'Singapore'),    ('Thailand', 'Bangkok'),    ('Vietnam', 'Hanoi')]Then, the code can proceed to get their coordinates via the Wikipedia API with the correct names.Fourth subtask (code): \\

\textbf{\textless begin\_call\_subtask\textgreater}
Write Python code to retrieve coordinates for each of the ASEAN capitals (using the corrected capital name for Myanmar as Naypyidaw) from Wikipedia, compute pairwise distances with Haversine formula, identify the maximum distance pair, and return them in alphabetical order. 
\textbf{\textless end\_call\_subtask\textgreater}\\

\textbf{\textless begin\_subtask\_result\textgreater} 
Result from Code-Agent:The code attempts to scrape coordinates for ASEAN capitals from Wikipedia, ensuring that Myanmar's capital is correctly listed as Naypyidaw. ...The maximum distance pair between ASEAN capitals is Jakarta and Naypyidaw, returned in alphabetical order.
\textbf{\textless end\_subtask\_result\textgreater} \\

So the user's answer should be boxed\{Indonesia, Myanmar\} \\
    \bottomrule
    \end{tabular}
    \label{tab:case}
\end{table*}

\end{document}